\documentclass[conference]{IEEEtran}
\IEEEoverridecommandlockouts
\usepackage{cite}
\usepackage{amsmath,amssymb,amsfonts}
\usepackage{algorithmic}
\usepackage{graphicx}
\usepackage{textcomp}
\usepackage{xcolor}
\def\BibTeX{{\rm B\kern-.05em{\sc i\kern-.025em b}\kern-.08em
    T\kern-.1667em\lower.7ex\hbox{E}\kern-.125emX}}
\begin{document}

\title{Aerodynamics and Sensing Analysis for Efficient Drone-Based Parcel Delivery \\}

\author{\IEEEauthorblockN{Avishkar Seth\textsuperscript{1}\IEEEauthorrefmark{1}, Alice  James\textsuperscript{1}, Endrowednes Kuantama\textsuperscript{2}, Subhas Mukhopadhyay\textsuperscript{1}, Richard Han\textsuperscript{2}}
\IEEEauthorblockA{\textit{School of Engineering, School of Computing } \\
\textit{Macquarie University}\\
Sydney, Australia \\
avishkar.seth@mq.edu.au}
\thanks{Article Published: doi: 10.1109/ICST59744.2023.10460847}

}
\maketitle


\begin{abstract}

In an era of rapid urbanization and e-commerce growth, efficient parcel delivery methods are crucial. This paper presents a detailed study of the aerodynamics and sensing analysis of drones for parcel delivery. Utilizing Computational Fluid Dynamics (CFD), the study offers a comprehensive airflow analysis, revealing the aerodynamic forces affecting drone stability due to payload capacity. A multidisciplinary approach is employed, integrating mechanical design, control theory, and sensing systems to address the complex issue of parcel positioning. The experimental validation section rigorously tests different size payloads and their positions and impact on drones with maximum thrusts of 2000 gf. The findings prove the drone's capacity to lift a large payload that covers up to 50 percent of the propeller, thereby contributing to optimizing drone designs and sustainable parcel delivery systems. It has been observed that the drone can lift a large payload smoothly when placed above the drone, with an error rate as low as 0.1 percent for roll, pitch, and yaw. This work paved the way for more versatile, real-world applications of drone technology, setting a new standard in the field.

\end{abstract}


\section{Introduction}
In the era of increasing urbanization and the exponential growth of e-commerce, efficient and sustainable parcel delivery methods have become a critical concern \cite{Eskandaripour2023Last-MileFuture}. Traditional ground-based delivery systems are facing challenges such as traffic congestion, increased carbon emissions, and inefficiencies in last-mile delivery. Companies like Amazon Prime Air and DHL Parcelcopter are pioneering drone deliveries, primarily in less populated areas . Drone deliveries offer a promising alternative to the rising issues of delivery as evidenced by a 2018 AlphaBeta report on the Australian Capital Territory \cite{Seth2023May}.

The majority of cargo drone prototypes currently in use are multicopters, which can take use of vertical takeoff and landing capabilities to ensure exact deliveries even in crowded areas \cite{shakhatreh2019unmanned, singireddy2018technology, bamburry2015drones}. These platforms, however, frequently require distinct take-off and landing slots and are built for long-range transportation. Additionally, the existing drone designs used by logistics firms have limitations when it comes to their ability to deliver packages of various sizes, especially those bigger than the drone itself. This restriction is due to the fact that packages are typically placed beneath the drone's propeller plane, which creates a lot of aerodynamic drag and lowers the drone's lifting capacity.

Nonetheless, the current design approach imposes constraints on the size of parcels that can be accommodated. A more compact drone design, capable of carrying oversized parcels relative to its rotor dimensions, could mitigate the need for such larger, less efficient platforms \cite{Saunders2023Jul, dissanayaka2023review}. Broadly classified, aerial robots are configured in many different ways with additional components while managing their CoG (Center of Gravity) \cite{MendozaMendoza2020SnakeAM}. Various design considerations need to be made while configuring the placement and positioning of loads on a UAV.

The type of sensors and peripherals used on board a UAV can vary depending on the application. A quadcopter's control system depends on orientation sensors like gyroscopes, accelerometers, magnetometers, and GPS to continually monitor its location and identify and resolve any movement issues using sensor fusion techniques \cite{nazarahari202140, schopp2009sensor, Mahony2012MultirotorQuadrotor}. Flight stability is critically important by the control system, which is visible as well as through sensor data logs that show acceleration and angle changes in three-axis coordinates \cite{ducard2021review, shi2020adaptive, du2012industrial}. Rotor speed monitoring is critical to improve sensor performance and reduce position error, especially during hovering and throttling phases \cite{jabeur2022optimized}. For real-time monitoring and control modifications, a variety of techniques can be used, including test bench systems, spherical models for UAVs, and cascade iterative algorithms \cite{Kuantama2018FlightInformation}. Most importantly, when creating a sensing mechanism for airflow analysis and downwash of a UAV, an anemometer is used to check wind speed below the propeller blades. When using an oversized parcel however, this wind slipstream is blocked causing reduced thrust. This problem can be analysed using anemometer, IMU sensors, and barometer for altitude sensing.

\par In this paper, a method for aerodynamics and sensing analysis for efficient drone-based parcel delivery is proposed. Building on our earlier work \cite{seth2022vertical} on parcel delivery, this paper focuses on the development of the UAV's different parcel positions and its effects on the airflow and flight thrust. This paper's main contributions are as follows:

\begin{itemize}
\item   \textbf{Airflow Analysis:} This paper introduces a comprehensive airflow analysis for a drone's parcel lift capability, focusing on the aerodynamic forces and their impact on drone stability and payload capacity. The study employs Computational Fluid Dynamics (CFD) to simulate the airflow patterns, providing valuable insights into optimizing drone designs for efficient parcel delivery.

\item   \textbf{Multi-disciplinary Approach to Parcel Positioning:} The paper addresses the complex issue of parcel positioning on drones by integrating mechanical design, control theory, and sensing systems. It explores the challenges related to the center of gravity and aerial manipulation, offering a holistic approach to overcoming these limitations.

\item \textbf{Experimental Validation and Payload Positioning:} The paper includes an experimental section that validates the theoretical findings. It tests different payload positions and their impact on drone stability and aerodynamics, thereby providing practical guidelines for drone-based parcel delivery systems. 

\end{itemize}

\section{Related Work}
Computational Fluid Dynamics (CFD) is a crucial tool in the study of drone aerodynamics for modelling airflow patterns around the drone's body and propellers. Aerodynamic forces including lift, drag, and thrust may be precisely modeled using CFD, which is essential for the drone's stability and mobility. As mentioned in the study \cite{kim2022aerodynamic}, CFD can specifically be used to optimize the angle of attack for drone wings to increase lift-to-drag ratios. Another work uses CFD to produce a data-driven model for forecasting lift and drag coefficients throughout a flapping cycle for flapping wing micro air vehicles \cite{Calado_2023}. These applications of CFD are instrumental in enhancing the performance and reliability of drones, particularly in complex flight conditions. In another study \cite{duan2023machine}, machine learning is used to improve real-time predictions of aerodynamic forces using sparse pressure sensor inputs from unmanned aerial vehicles. This increases the forecasts' speed and accuracy.

\par Aerodynamic analysis and disturbances have been closely studied in the literature in the past decade. A close related research topic presented in 2020 \cite{Kornatowski2020DownsideDelivery, Kornatowski2017AnDrone, kornatowski2020morphing}, discusses an approach of placing a parcel above the drone. The paper's major limitations lie in its focus on hovering flight, neglecting the complexities of cruise flight and real-world delivery scenarios. The work in \cite{Kornatowski2017AnDrone} presents an origami-based parcel delivery drone, and in \cite{kornatowski2020morphing}, the authors present a similar enclosed close proximity parcel delivery drone. Resilience to collisions is another important aspect of drone design. Bioinspired strategies, such as insect-inspired mechanical resilience, can be employed to design more resilient drones to collisions \cite{Mintchev2017}. This can help mitigate the vulnerability of drones to collisions caused by pilot mistakes or system failures \cite{Mintchev2017}.

\par In summary, while existing research has made strides in aerodynamics and drone design, it often falls short in addressing the full spectrum of real-world applications. The studies, for instance, focus primarily on hovering flight, overlooking the nuances of cruise flight and actual delivery conditions. Origami-based and enclosed designs offer innovative solutions but still operate within a limited scope. Our approach, by contrast, builds on this research through safety and aerodynamic efficiency, tackling both take-off and cruise flight conditions. It advances the field and opens the door for more versatile, real-world applications of drone technology.

\section{System Methodology}

In this study, the challenge of placing a parcel above is an aerodynamics and sensing integrated issue. The objective of this system is to enable a quadcopter to maneuver through using different oversized parcels above the central section. To verify the theory, three configurations and sizes of UAVs are employed as shown in Table~\ref{tab:specs}. The table outlines the specifications of three different drones, focusing on four key attributes: size, frame material, motor power, and maximum load capacity. 

\begin{table} [ht]
  \caption{Three Varying Drone Dimensions and Specifications}
  \label{tab:specs}
   \resizebox{\columnwidth}{!}
    {\begin{tabular}{|c|c|c|c|} 
    \hline
     \textbf{Drone Size (mm)} & \textbf{Frame material} & \textbf{Motor} & \textbf{Max Load}\\ \hline
     295 x 295 x 55 & Carbon Fiber & 1750 kV & 1100 gms \\ \hline
     450 x 450 x 55 & Polyamide-Nylon & 930 kV & 2280 gms \\ \hline
     675 x 675 x 210 & Carbon Fiber & 400 kV & 3200 gms \\ \hline
  \end{tabular}}
\end{table}

The drones vary in size, from a compact 295 x 295 x 55 mm model to a more substantial 675 x 675 x 210 mm version. Frame materials include Carbon Fiber, prized for its lightweight strength, and polyamide nylon, which offers robustness at a slightly higher weight. Motor power, expressed in kV, reveals a trade-off between speed and torque. Finally, the maximum load capacity ranges from 1100 grams for the smallest drone to 3200 grams for the largest. This diverse set of UAVs allows us to rigorously test our approach across various scenarios, ensuring its robustness and applicability.

\begin{figure}[ht]    
    \centering
    \includegraphics[width=1\linewidth]{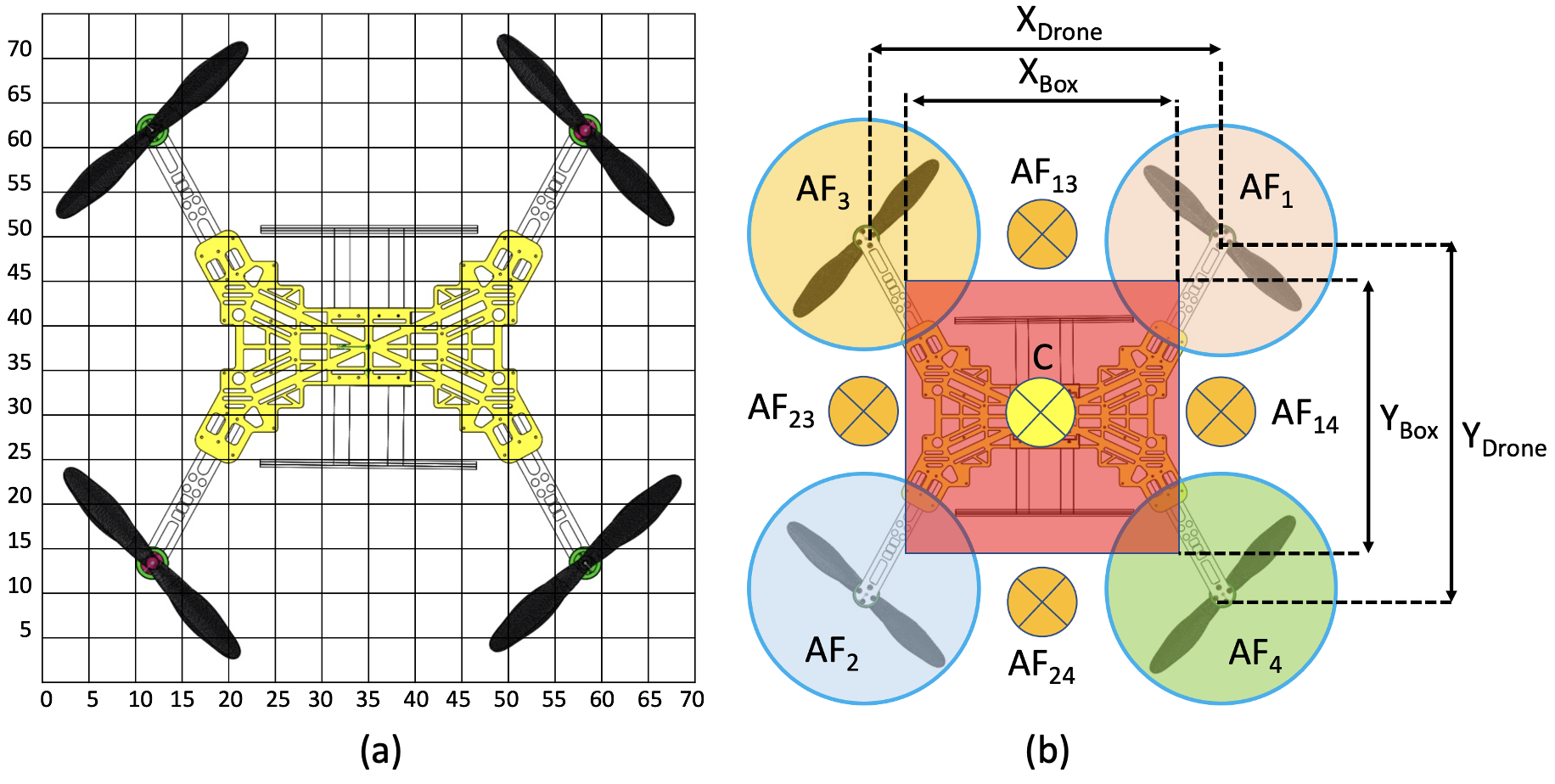}
	\caption{The schematic representation of the geometric relationship between the dimensions of the box placed above the drone and the eight airflow variations on and around the propeller slipstream}
\label{drone_dimensions}  
\end{figure}

Figure~\ref{drone_dimensions} shows the schematic representation of the geometric relationship between the dimensions of the box placed above (denoted by \( X_{\text{Box}} \), \( Y_{\text{Box}} \)) the drone (denoted by \( X_{\text{Drone}} \), \( Y_{\text{Drone}} \)) and the eight airflow variations on (denoted by \( AF_1 \) to \( AF_4 \)) and around (\( AF_{13} \), \( AF_{14} \), \( AF_{23} \), \( AF_{24} \)) the propeller slipstream. This study is done to calculate the appropriate size of the box with respect to the drone's dimensions. It is crucial to maintain the centre of gravity by placing the parcel in the centre. The aerodynamics airflow is tested at the (AF) sections while in hover mode using an anemometer placed below the propeller slipstreams. The compact design also enables the drone to navigate through cluttered or hard-to-reach areas, which are typically inaccessible for drones with parcels placed between the propellers. Moreover, this configuration minimizes the aerodynamic interference of the parcel, allowing for more efficient flight dynamics. 

\section{Airflow Analysis}

\begin{figure}[htpb!]    
    \centering
    \includegraphics[width=1\linewidth]{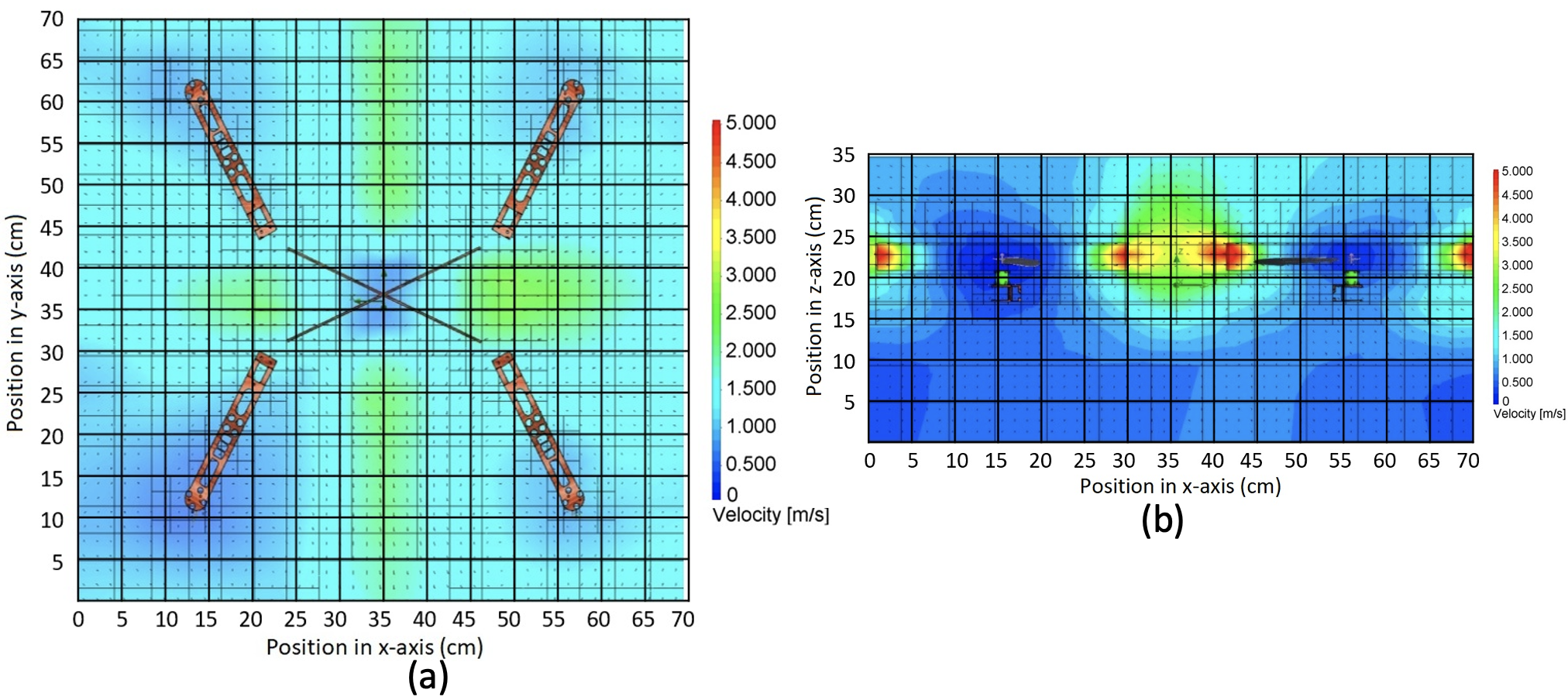}
	\caption{CFD simulation images of the UAV when no payload is placed depicting the Velocity [m/s] of the propellers (a) Bottom View (b) Side View}
\label{cfd_1}  
\end{figure}

The data analyzing airflow distribution is verified while the drone remains stationary in a hovering state at each analyzed point. The collected data exhibited significant sensitivity to the air patterns created by the propeller's vortex field and the prevailing wind conditions. We applied the vortex method, as described in \cite{kuantama2019design}, to scrutinize the aerodynamic performance of the aircraft. Earlier research by the authors in \cite{kuantama2019design} provides detailed insights into the design of the drone's propellers and the analysis of its flight stability.
We conducted a computational fluid dynamics (CFD) analysis using SolidWorks software to determine the total thrust generated. 
Each propeller was situated within its designated rotation area, where a pair of propellers diagonally across from each other rotated in a clockwise (CW) direction, while the other diagonal pair rotated counterclockwise (CCW) at velocities ranging from 0 to maximum rpm for each of the three drones. This simulation yielded data on the vertical force produced and the total air velocity within all propeller rotation areas.
For the drone with 13-inch propellers, each rotor had the capability to generate a total velocity within 5000 m/s. The vortex field was also simulated using the same software and parameters. Figure ~\ref{cfd_1} below visually represents the drone design to lift the payload.

\par The transportation of the large-sized payload represents a critical process essential for several applications, and this process can be susceptible to disturbances caused by the vortex generated by the propeller. Factors such as the payload's location and wind resistance significantly affect the measurement outcomes. When determining the placement of the payload, one must carefully consider the drone's center of gravity while also ensuring the quadcopter's stability due to airflow turbulence.

While it may seem tempting to use an extended pole below the drone to position the payload outside the propeller's vortex field, this approach is not advisable as it can potentially compromise the stability of the drone and the payload itself due to the pendulum effect. Although it may have a limited impact when using a small payload, attaching an extended pole carrying a larger is more susceptible to the influence of wind disturbances. Consequently, the design and computational analysis for placing the payload on the quadcopter's frame was restricted to two specific points: above the frame or closely attached below the frame.

\par The second option involves placing the payload at the bottom of the main frame, ensuring the payload is unaffected by the propeller's vortex; however, that restricts the size capacity of the payload. On the other hand, placing the payload above corresponds to a larger size limit, causing the propeller's vortex to remain unaffected. Before conducting flight tests, both positions were thoroughly analyzed using computational fluid dynamics (CFD).
\par  Drones use spinning propellers to move and create forces to fly. These forces are thrust, drag, lift, and weight, affecting how the drone moves. Thrust is like a push from the propellers that helps the drone go up and opposes the pull of gravity, trying to bring it down. Lift and thrust together impact which way the drone goes, and lift depends on how fast the air is moving, how thick the air is, the size of the wings, and how they're angled. Drag is like air resistance, slowing the drone down as it moves forward. Weight is the force of gravity pulling the drone down. All of these forces work together to determine how the drone flies, and they can change depending on how fast the air is moving and the shape of the propellers. The location of the payload relative to the drone's center of gravity significantly affects these forces—thrust, drag, lift, and weight. Optimal payload placement, closer to the center of gravity, facilitates a balanced weight distribution and minimizes the strain on thrust requirements. Conversely, an off-center or poorly positioned payload can disrupt equilibrium, necessitating increased thrust to compensate for weight discrepancies. Furthermore, the payload's spatial configuration is crucial in determining the drone's aerodynamic performance, with streamlined payload arrangements reducing drag resistance. Hence, meticulous consideration of payload positioning is paramount for efficient force management and maintaining flight stability and control. 
Air resistance acts as a frictional force that impedes motion. In this study, if the wind blows vertically due to the payload's position and the drone is in a hover state, turbulence will occur and cause instability. 
\begin{figure}[htpb!]    
    \centering
    \includegraphics[width=1\linewidth]{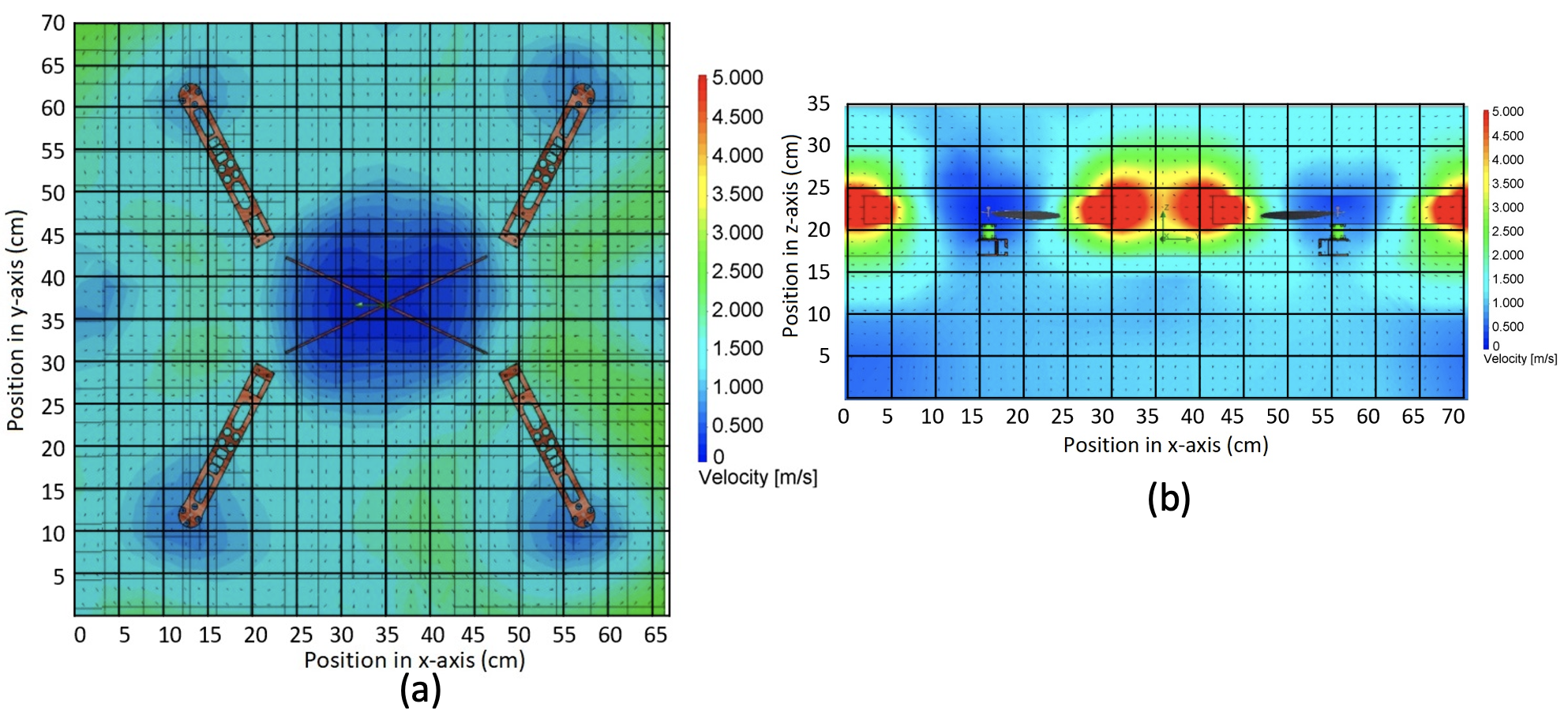}
	\caption{CFD simulation images of the UAV when a payload is placed depicting a higher aerodynamic disturbance sensed around the propellers (a) Bottom View (b) Side View}
\label{cfd_2}  
\end{figure}

Wind effects were systematically examined under constant wind velocity, assessing the aerodynamic force generated by relative wind on the propeller. Computational analysis of lift and drag coefficients identified the airfoil extending along the propeller's radius, with the distribution of drag (\( C_{\text{Drag}} \)) and lift (\( C_{\text{Lift}} \)) coefficients scrutinized at the design point. These coefficients were determined using the Rayleigh equation, as represented in equations (1) and (2).

\begin{equation}
C_{\text{Drag}} =  \frac{2 F_{\text{Drag}}}{A_p \rho v^2}\
\end{equation}

\begin{equation}
C_{\text{Lift}} =  \frac{2 F_{\text{Lift}}}{A_p \rho v^2}\
\end{equation}

\par The drag and lift coefficients are contingent on several factors, including the propeller's surface area ($A_p$), the airflow surrounding the propeller \textit{(v)}, and the density of the fluid, which, in this instance, is air with a density ($\rho$) of 1.225 kg/m³. These coefficients are integral to understanding the forces in both drag (\( F_{\text{Drag}} \)) and lift (\( F_{\text{Lift}} \)) conditions. This formula was applied to determine the drag and lift coefficients for the tested propeller. Figure ~\ref{cfd_2} illustrates the airflow patterns around the propeller at a specific airflow velocity.  The spinning of the propellers initiates a relative wind, which is closely tied to the aerodynamic motion of the drone. This airflow produces relative wind with flow vectors pointing downward. By effectively adjusting the payload's position and maintaining each propeller's angular velocity, it becomes possible to stabilize the drone's flight.

\section{Payload Position Validation}
The impact of payload position on the propeller influences drag and lift forces. As perceived within the drone's body frame, wind direction is denoted by ($\psi$) for yaw angle changes and ($\theta$) for pitch angle changes. To describe the orientation of the moving frame concerning the fixed frame, angular position vectors were determined using a kinematic moving frame theorem. External effects from the wind that affect the drones are quantified through lift and drag forces in an inertial frame, as shown in equation (3-5).

\begin{equation}
F_{\text{pitch}} =  - F_{\text{Drag}} \cos \theta \cos \psi - ( F_{\text{Lift}} - mg) \sin \theta \cos \psi
\end{equation}
\begin{equation}
F_{\text{roll}} =  - F_{\text{Drag}} \sin \theta + ( F_{\text{Lift}} - mg) \cos \theta + T
\end{equation}
\begin{equation}
F_{\text{yaw}} =  - F_{\text{Drag}} \cos \theta \sin \psi - ( F_{\text{Lift}} - mg) \sin \theta \sin \psi
\end{equation}

\begin{figure}[htpb!]    
    \centering
    \includegraphics[width=1\linewidth]{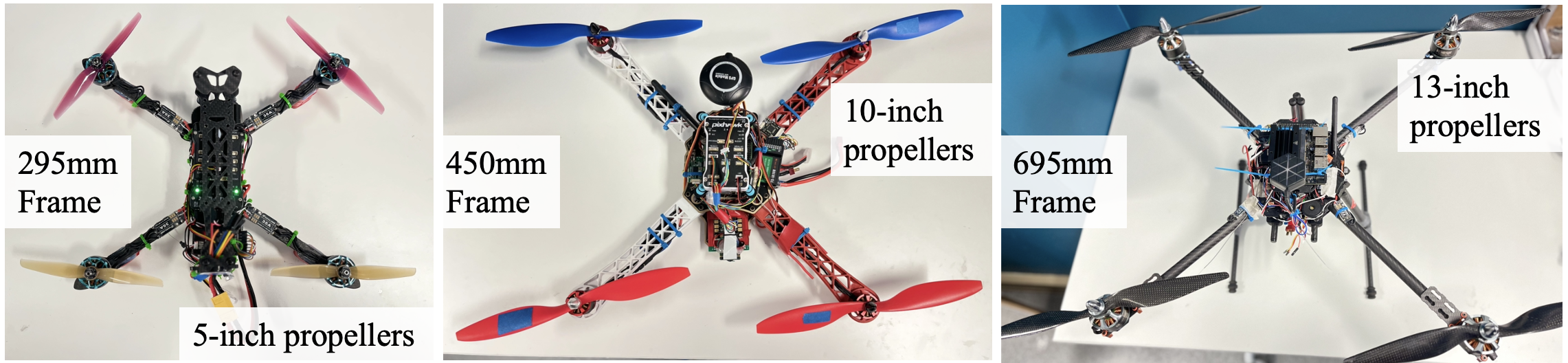}
	\caption{Photographs of the three different drones with varying dimensions and configurations}
\label{diff_drone}  
\end{figure}

By regulating the amount of thrust, i.e., drag and lift force, the drone might be kept in a hovering posture while maintaining the required height and attitude. As discussed previously, three different sizes of drones were used in this research, as seen in Figure~\ref{diff_drone}. The tests were conducted to validate the optimal position of the payload if the dimensions were larger than the propeller.

\begin{figure}[htpb!]    
    \centering
    \includegraphics[width=1\linewidth]{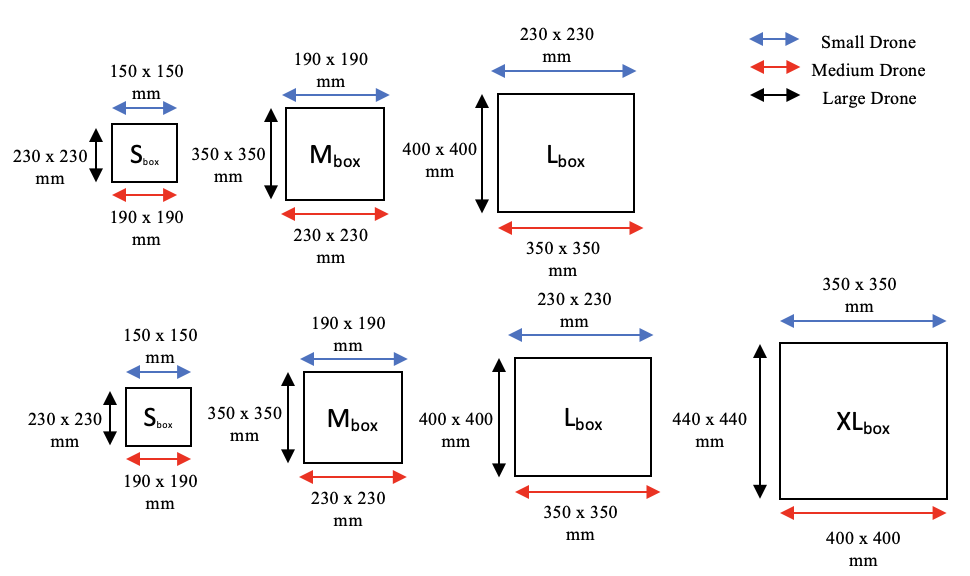}
	\caption{Dimensions of different sized boxes that are tested on the drones ranging from small to large}
\label{parcel_dim}  
\end{figure}

Figure~\ref{parcel_dim}  illustrates the dimensions of the different size payloads tested. The top row shows the three sizes of boxes that were deployed below the drone, and the bottom row gives the dimensions of the four different sizes of payloads placed above the drone.

In the experimental setup, the drones were positioned in an altitude-hold position, where different parcel payloads were mounted individually either below or above the drone. The anemometer measured the inflow and outflow of air below the propeller and between the propeller and the payload. Utilizing air structure analysis, we could dynamically observe how the drone's frame was affected within this payload mount setup. To maintain a stable hovering position, the drone had to continuously adapt to the influence of the payload,  which was analyzed to verify the optimal position for the payload; the results are discussed in the following section. 

\section{Experimental Evaluation}
\subsection{Airlfow  tests and Balancing Validation}
\begin{figure}[htpb!]    
    \centering
    \includegraphics[width=1\linewidth]{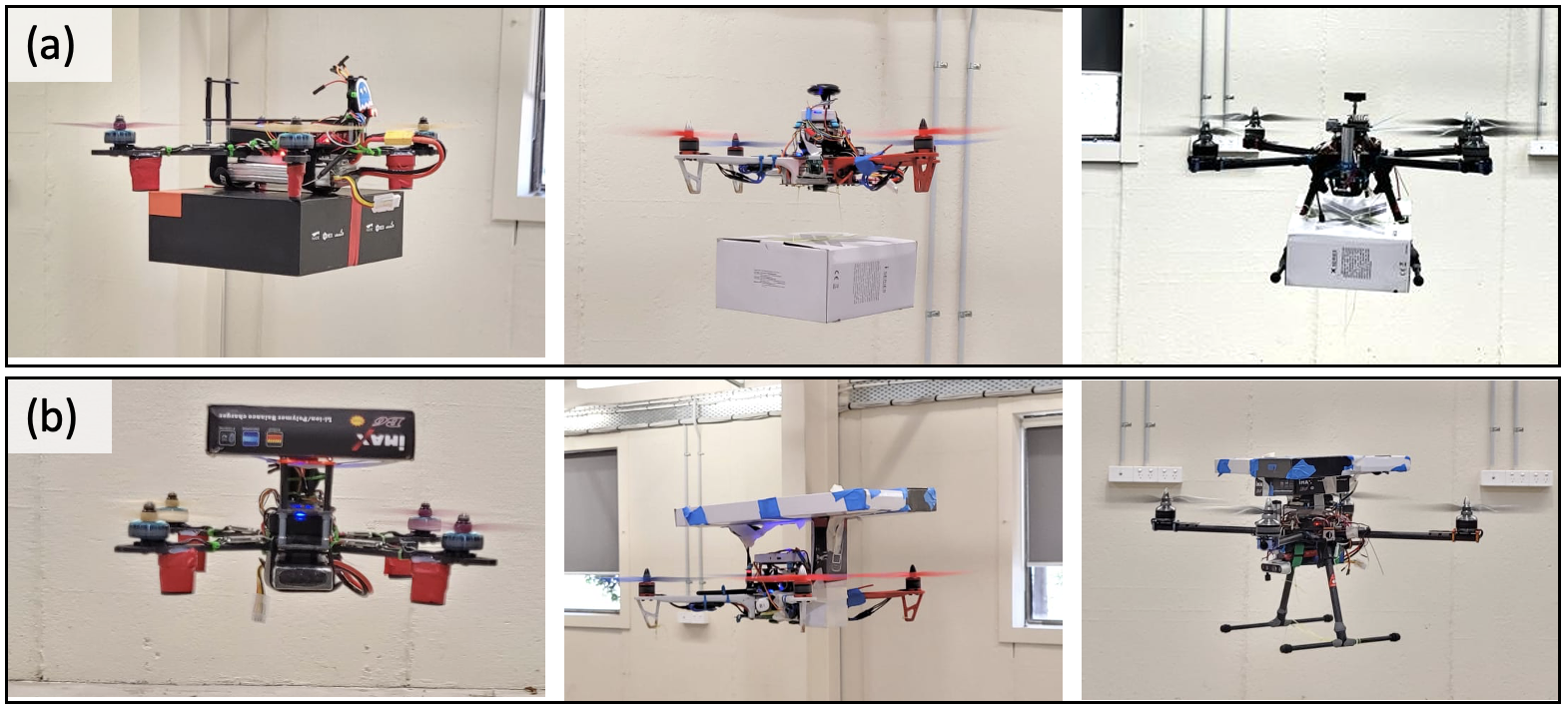}
	\caption{Photographs of the three different drone sizes carrying payloads (a) below and (b) above the drone while hovering}
\label{3_drone_parcel}  
\end{figure}

For the experimental results of placing packages on delivery drones, we wanted to see how it affects the drone's airflow distribution and how stable the drone flight is. Figure~\ref{3_drone_parcel} shows the three drones mid-flight with a payload, (a) shows the payload positioned below, and (b) shows the payload positioned above. An important result is presented: where you put the package can make a big difference in the airflow distribution when the drone is flying.

\begin{figure}[htpb!]    
    \centering
    \includegraphics[width=1\linewidth]{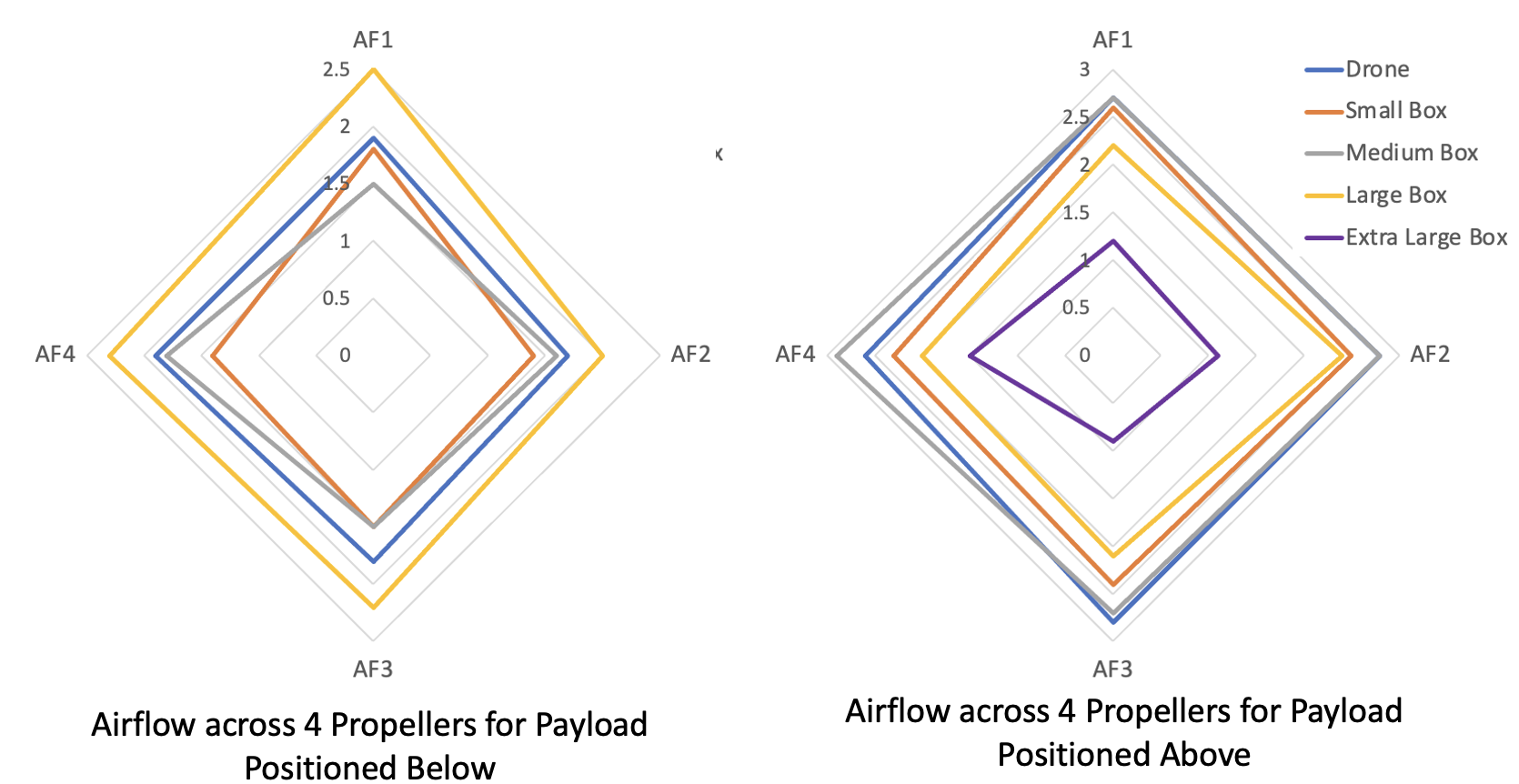}
	\caption{The Radar Chart of airflow sensing analysis across propellers of the quadcopter for payload positioned below (left image) or above (right image) using real-world experiment values}
\label{airflowtest}  
\end{figure}

When the payload encompasses the drone's spinning propellers, we notice that the turbulence increases, making the drone unstable. The same is observed in Figure~\ref{airflowtest} (a), which shows turbulence for the large payload positioned below the drone.
Observing Figure~\ref{airflowtest}, it is seen that putting a bigger package under the drone causes even more vibration. This outcome underscores the significance of payload positioning in mitigating turbulence-related challenges during drone flights. The data obtained in this analysis highlight the importance of carefully considering payload placement to maintain stable drone operations. 

\begin{figure}[htpb!]    
    \centering
    \includegraphics[width=1\linewidth]{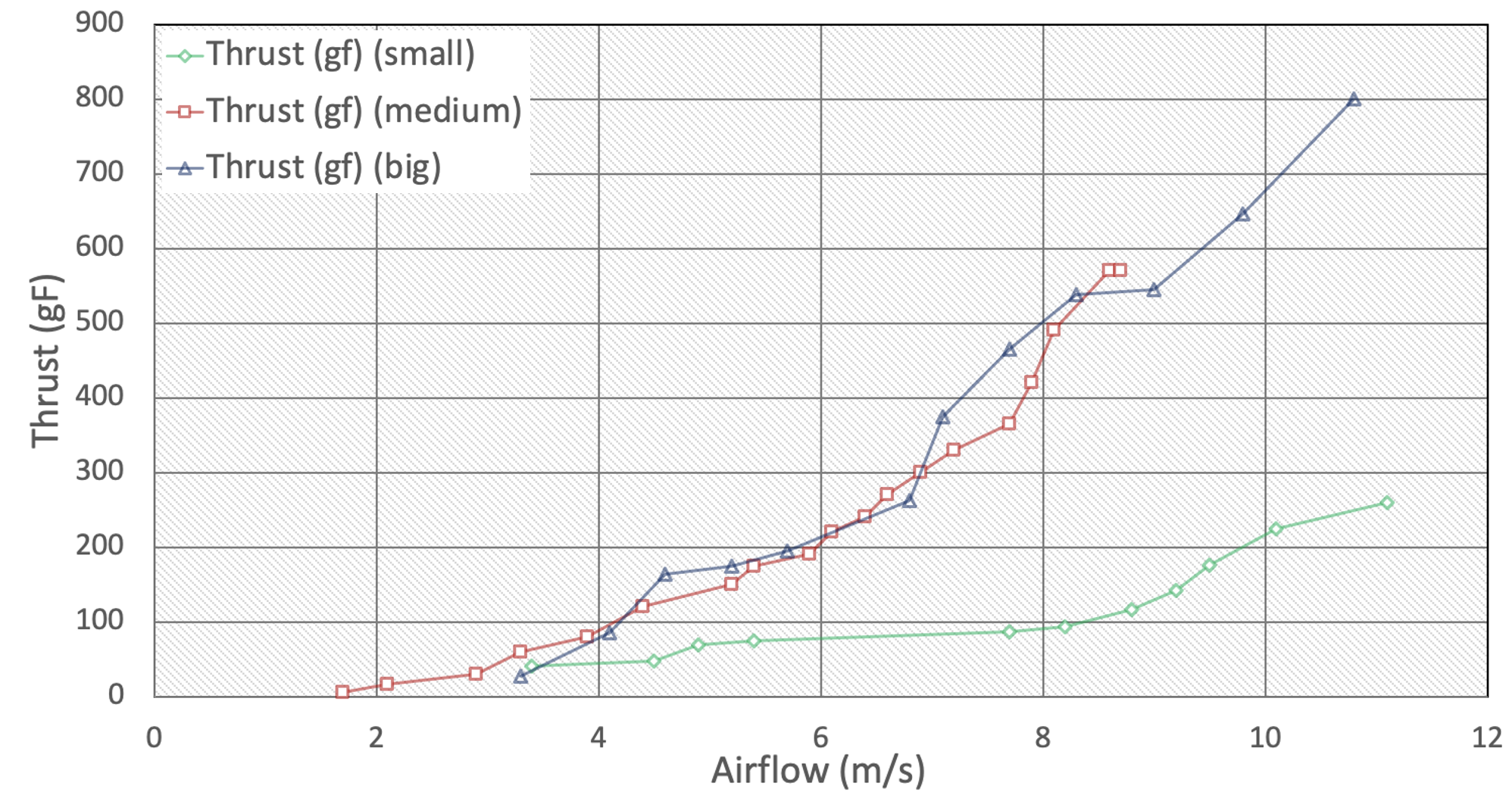}
	\caption{Graphical representation of thrust verification values for three different sizes of drones with payload position below}
\label{thrustvsAF}  
\end{figure}

When the propeller spins at the maximum rpm, its thrust increases notably. A single rotor can produce a maximum thrust range of 1000 to 2000 grams-force (gf) for the three drones tested. This means that when all four rotors work together, the drone can generate a total thrust of 8 kilograms-force (kgf). This information helps us estimate how much weight the drone can carry and how fast it can fly. A comparison is illustrated to ensure our experimental results are reliable, highlighting the change in thrust due to the disturbance in the airflow. Figure~\ref{thrustvsAF} compares our experimental data for the small,  medium, and big drones with the below payload.

The results indicate that our simulation data closely match the specifications of experimental data rotors, as the thrust measurements are significantly lowered due to the payload position below the drone. It is observed that there is no significant change in thrust when the payload is positioned above. Thus, a large payload with up to 50 percent propeller blade coverage can be accommodated above the drone with negligible turbulence compared to the existing method of lifting a package from below.

\subsection{Flight tests}

\begin{figure}[htpb!]    
    \centering
    \includegraphics[width=0.9\linewidth]{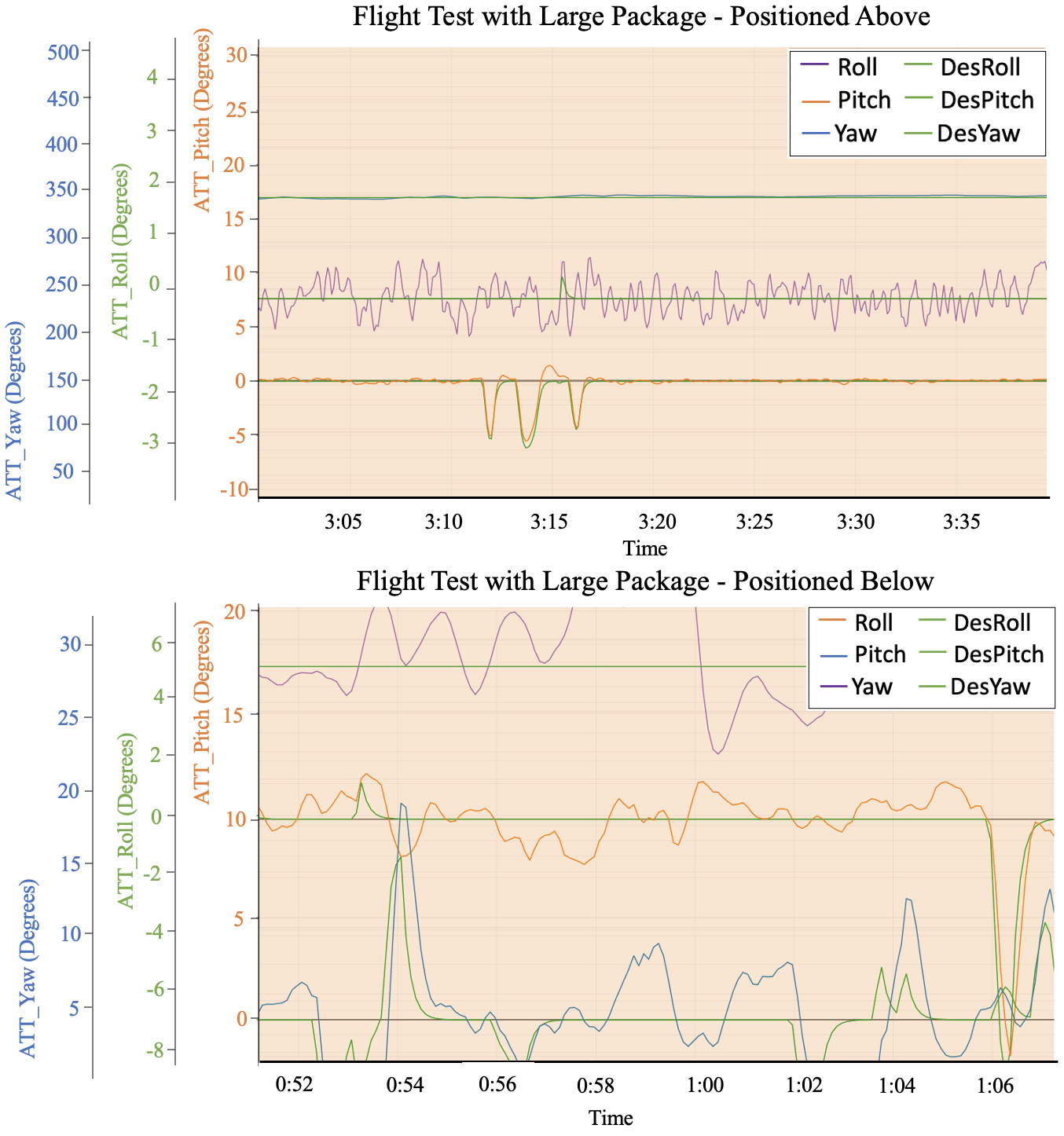}
	\caption{IMU sensor values obtained from flight tests used to verify accurate desired vs actual Roll Pitch Yaw values of the Drone while carrying payloads above or below the drone}
\label{RPY-stability}  
\end{figure}

The proposed design ensures that the CoG is positioned close to the propeller plane the payload irrespective of the payload presence. Flight trials were conducted using the quadcopter to assess the feasibility of positioning a package above the rotor plane while minimally impacting lift. During the test, the quadcopter maintained a 2.5-meter altitude using 55\% throttle to handle a 200g payload, guided by a downward facing ranging sensor to maintain altitude. The drone operated on basic Ardupilot flight firmware, designed for an X-configured quadcopter, which is less intricate than controllers used for long-tether parcel delivery. A pilot ensured stable positioning throughout the experiment. Figure~\ref{RPY-stability} shows the IMU sensor values obtained from flight tests used to verify accurate desired vs actual Roll Pitch Yaw values of the Drone while carrying payloads above or below the drone. The lines in green represent the desired values of Roll, Pitch, and Yaw, respectively, from the top. The error rate for roll and pitch is very high (about 20\%) when the oversized parcel is placed below the drone. However, the error rate for roll, pitch and yaw is approximately 0.1\% proving the drone can maintain its hover state stability while a parcel is placed above.

\section{Conclusion}
This study has comprehensively analyzed aerodynamics and sensing mechanisms for efficient drone-based parcel delivery. Utilizing Computational Fluid Dynamics (CFD) simulations and experimental validations, the research has addressed the critical issue of payload positioning and its impact on drone stability and aerodynamics. 
The study also employed a multidisciplinary approach, integrating mechanical design, control theory, and sensing systems to optimize payload positioning.
For thrust ranges between 1000 - 2000 gf, the error rates for roll, pitch, and yaw were as low as 0.1\% when the payload was positioned above the drone. This starkly contrasts a 20\% error rate observed when the payload was positioned below the drone, thereby validating the aerodynamic efficiency of the above-the-drone payload positioning.
The experimental results corroborated the CFD simulations, showing a close match between the simulated and actual thrust values. The maximum achievable payload size can cover up to 50\% of the propeller blade, which provides valuable insights and better opportunities into the payload capacities of drones of varying sizes. In summary, the findings of this study offer invaluable guidelines for optimizing drone designs for parcel delivery, setting a new standard in the field. Future work should focus on real-world applications, including but not limited to, varying weather conditions, dynamic payload positioning, and integration with existing logistics systems.

\section*{Acknowledgment}

The authors acknowledge the continued support from Macquarie University through the Computing and Engineering Faculty for providing the resources and space required for this research.\

\bibliographystyle{IEEEtran}
\bibliography{references_airflow}

\end{document}